\documentclass[preprint,12pt]{elsarticle}

\usepackage{amssymb}
\usepackage{amsmath}
\usepackage{siunitx}

\journal{Solar Energy}

\begin{document}

\begin{frontmatter}



\title{Lightweight Transformer-Driven Segmentation of Hotspots and Snail Trails in Solar PV Thermal Imagery}


\author[1,2]{Deepak Joshi} 


\author[1]{Mayukha Pal\corref{mycorrespondingauthor}}
\ead{mayukha.pal@in.abb.com}
\cortext[mycorrespondingauthor]{Corresponding author}

\affiliation[1]{organization={ABB Ability Innovation Center},
     addressline={Asea Brown Boveri Company},
    city={Hyderabad},
    postcode={500084},
    state={Telangana},
    country={India}}    

\affiliation[2]{organization={Department of Electrical Engineering},
    addressline={Indian Institute of Technology},
    city={Hyderabad},
    postcode={502285},
    state={Telangana},
    country={India}}

\begin{abstract}

Accurate detection of defects such as hotspots and snail trails in photovoltaic (PV) modules is essential for maintaining energy efficiency and system reliability. This work presents a supervised deep learning framework for segmenting thermal infrared images of PV panels, using a dataset of 277 aerial thermographic images captured by a Zenmuse XT infrared camera mounted on a DJI Matrice 100 drone. The preprocessing pipeline includes image resizing, CLAHE-based contrast enhancement, denoising, and normalisation. A lightweight semantic segmentation model based on SegFormer is developed, featuring a customised Transformer encoder and streamlined decoder, and fine-tuned on annotated images with manually labelled defect regions. To evaluate performance, we benchmark our model against U-Net, DeepLabV3, PSPNet, and Mask2Former using consistent preprocessing and augmentation. Evaluation metrics include per-class Dice score, F1-score, Cohen’s kappa, mean IoU, and pixel accuracy. The SegFormer-based model outperforms baselines in accuracy and efficiency, particularly for segmenting small and irregular defects. Its lightweight design enables real-time deployment on edge devices and seamless integration with drone-based systems for automated inspection of large-scale solar farms.

\end{abstract}

\begin{graphicalabstract}
\begin{figure*}[ht!]
\centering
	\includegraphics[width=12cm,height=4.5cm]{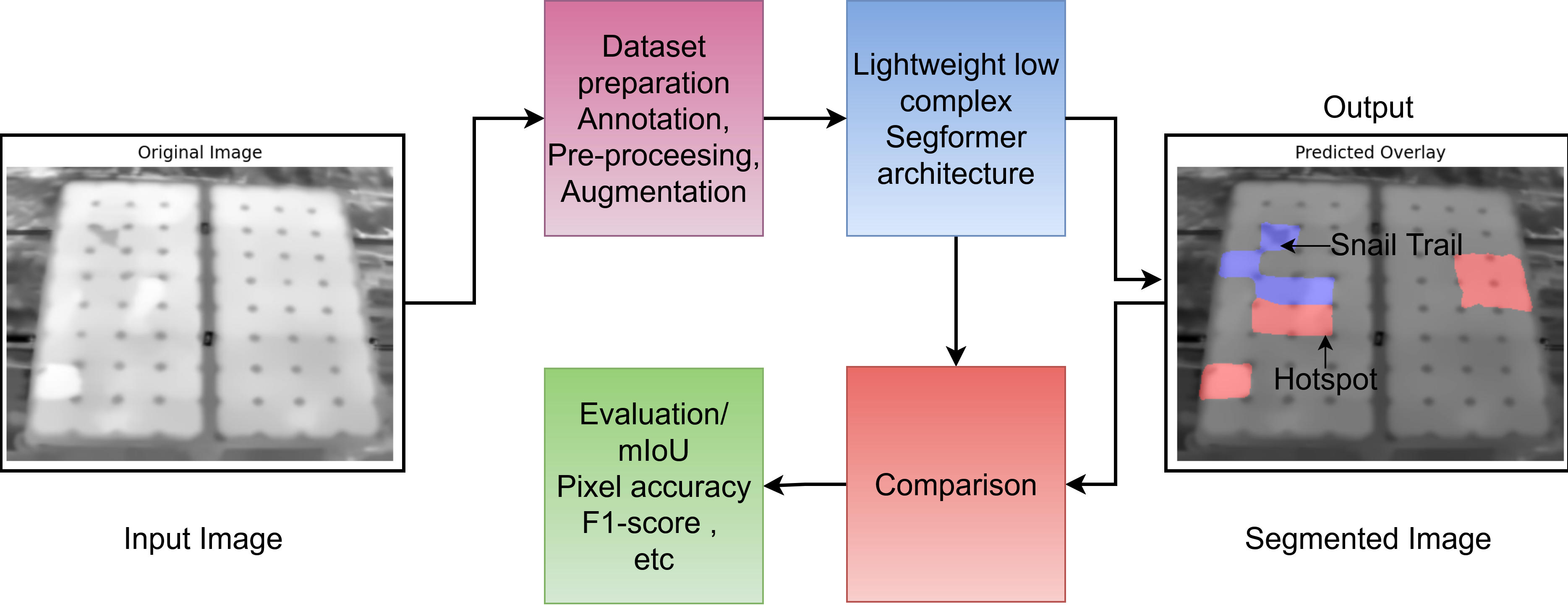}
	
\end{figure*}
\end{graphicalabstract}

\begin{highlights}
\item A lightweight and customized SegFormer-based segmentation framework is developed for the analysis of thermal images in solar photovoltaic (PV) modules.
\item The proposed model is rigorously evaluated using a suite of metrics, including mean Intersection over Union (mIoU), pixel-wise accuracy, F1-score, Dice coefficient, and Cohen’s kappa score.
\item Detailed class-wise segmentation performance is reported, with a specific focus on identifying hotspot and snail trail defects.
\item The framework's effectiveness is further validated through comparative experiments against established segmentation architectures such as U-Net, DeepLabV3, and PSPNet.
\end{highlights}


\end{frontmatter}




\section{Introduction}
The global push toward sustainable energy has accelerated in response to climate change, energy security concerns, and the environmental degradation caused by the prolonged use of fossil fuels. Traditional energy sources such as coal, oil, and natural gas are not only finite but are also major contributors to greenhouse gas emissions, which are driving global warming and extreme climate events \cite{p1,p2}. In line with international agreements such as the Paris Climate Accord, countries worldwide are investing heavily in renewable energy sources to reduce carbon emissions and ensure a sustainable energy future. Among these, solar energy has emerged as one of the most promising and widely adopted alternatives due to its abundance, scalability, and zero-emission nature during operation \cite{p3}.

Photovoltaic (PV) technology, which directly converts solar radiation into electricity, has seen unprecedented growth in the last decade. According to recent estimates, global PV capacity surpassed 2.2 terawatts by the end of 2024 and is projected to continue expanding rapidly. Solar PV modules are now deployed in residential rooftops, commercial installations, and utility-scale solar farms across the globe \cite{p4}. 
However, the long-term efficiency and safety of PV systems are compromised by various types of physical and thermal defects such as hotspots, snail trails, microcracks, and delamination. These faults can emerge due to manufacturing imperfections, mechanical stress, or prolonged environmental exposure, and may lead to power loss, thermal degradation, and even safety hazards if left undetected \cite{p5}.

Among the commonly observed faults, hotspots and snail trails are particularly critical due to their impact on performance and their prevalence in real-world deployments. A hotspot forms when a portion of a PV cell experiences higher resistance—often due to shading, cell cracks, or soiling—which leads to localized heating as current is forced through the impaired region. This not only reduces the power output but also accelerates material degradation and can even result in module failure if not addressed \cite{p6}.

On the other hand, snail trails appear as discolored, worm-like patterns on the surface of PV modules. These are typically caused by microcracks within the silicon cells, often induced by mechanical stresses during transportation, installation, or operation. Over time, moisture ingress and chemical reactions with silver paste traces at these cracks lead to visible discoloration. While snail trails may initially seem superficial, they often signal underlying structural defects that contribute to long-term efficiency losses and reliability concerns in PV systems \cite{p7}.

Spataru et al. \cite{p8} introduced a two-dimensional matched filtering technique to localize micro-cracks in photovoltaic (PV) modules. Their method generates a binary location map by applying post-processing on filtered outputs, allowing for spatial identification of potential crack sites. Similarly, Rashmi et al. \cite{p9} leveraged the Mamdani fuzzy logic algorithm to extract micro-cracked areas and quantify their surface coverage in PV cell imagery, enabling area-based defect discrimination. Although both approaches offer targeted crack localization, they rely on classical image processing heuristics that are sensitive to parameter tuning. As noted by Jiang et al. \cite{p10}, such techniques depend heavily on expert-defined configurations for denoising and feature extraction. This manual dependency restricts their adaptability when operating under diverse or unpredictable imaging conditions, especially in the presence of complex backgrounds and varying textures.

Traditional image processing techniques have served as the foundation for early photovoltaic (PV) module defect analysis. Methods such as thresholding, edge detection, region growing, clustering, and graph-based segmentation have been widely adopted to detect surface anomalies. For example, S. Gayathri et al.\cite{p11} applied a hybrid Otsu-Canny edge detection algorithm to isolate micro-cracks in silicon solar cell images, and Montanez et al. \cite{p12} proposed a thermal analysis tool that segments individual PV modules from UAV-acquired infrared images using an automatic pipeline combining Otsu thresholding and contour detection to compute temperature statistics for fault identification. In a similar vein,  Prasshanth et al. \cite{p13} proposed another UAV-based study combined histogram-based statistical features (mean, variance, skewness), texture features derived from Gray-Level Co-occurrence Matrix (GLCM), and frequency-domain features using Discrete Wavelet Transform (DWT) to characterize thermal anomalies, followed by Rough Set Theory (RST) for classification and rule generation, offering a multi-feature perspective for PV fault analysis. Cao et al. \cite{p14} enhanced morphological and edge-based techniques for improved precision. Attia et al. \cite{p15} leveraged K-means clustering with hyperspectral imaging for real-time PV fault detection, and Zhang et al. \cite{p16} used a combination of Prewitt and Canny filters to identify micro-cracks. Despite their early promise, these techniques often exhibit poor generalization under complex environmental conditions and diverse imaging scenarios, limiting their robustness in real-world deployments.

To improve scalability and coverage in PV diagnostics, aerial inspection using UAVs has emerged as a practical solution. Quater et al. \cite{p17} proposed a UAV-based monitoring framework utilizing both thermal and optical imaging to detect a variety of module faults across solar farms.In a related effort, Chris et al. \cite{p18} implemented an automated detection system combining drone-captured thermal images with binarization and contour extraction to isolate deteriorated PV modules. Their system leveraged bounding box localization and integrated color-based analysis to enhance visual differentiation of damaged regions. Expanding on this direction, Aghaei et al.\cite{p19} developed a thermographic image analysis algorithm involving grayscale conversion, Gaussian filtering, binary segmentation, and Laplacian-based boundary modeling to highlight fault regions. Their subsequent studies \cite{p20,p21} investigated how flight altitude impacts defect visibility and computed degradation percentages across faulty regions. While effective in identifying thermal anomalies, these methods often lack fine-grained spatial resolution and fall short in delivering pixel-level precision needed for localized maintenance planning. Advancements in thermal analysis pipelines have further refined UAV-enabled PV inspection. Dotenco et al.\cite{p22} introduced a robust image-processing workflow to isolate individual PV modules and classify them into overheated modules, hotspots, and overheated substrings using thermal imagery. Complementing this, Lee et al.\cite{p23} developed a drone-assisted platform combining thermal and RGB images for enhanced fault detection in PV arrays. Although these frameworks demonstrate improvements in module-level diagnostics, they still do not achieve the pixel-wise segmentation fidelity essential for high-resolution fault localization and defect pattern characterization.

In recent years, deep learning has witnessed significant advancements, proving highly effective for complex pattern recognition across varied datasets. This progress has been particularly impactful in the domain of defect detection, where deep neural networks (DNNs) have been successfully applied for classification, localization, characterization, fault diagnosis, and monitoring, as evidenced by several studies in the literature \cite{p24,p25}. Building on this trend, recent works have explored tailored segmentation strategies for PV cell defect detection. For instance, a DSConv-enhanced encoder integrated with Efficient Channel Attention (ECA) was proposed to improve segmentation accuracy of microcracks in electroluminescence (EL) polarization images, demonstrating better feature representation with reduced computation \cite{p26}. Similarly, an attention-based dual-branch network was introduced to jointly perform classification and segmentation of micro-crack anomalies, leveraging multi-scale feature fusion and adaptive attention mechanisms to boost detection reliability in complex imaging scenarios \cite{p27}. Complementing these efforts, a unified deep learning approach based on the YOLOv5 architecture was presented for real-time surface damage detection across multiple renewable energy domains, including PV systems, wind turbines, and hydropower stations \cite{p28}. While this framework excels in bounding-box-based anomaly localization and supports cross-domain generalization, it lacks pixel-level precision and is not tailored for thermal data or resource-constrained edge deployment. Another noteworthy contribution employed transfer learning with EfficientNet-B0 to classify thermal images of PV modules as healthy or faulty. By fine-tuning the pretrained model and using data augmentation, this approach achieved 93.93\% classification accuracy, highlighting its potential for lightweight and accurate thermal fault detection \cite{p29}. Further advancing UAV-assisted diagnostics, Tang et al. \cite{p30} proposed a method combining histogram equalization and Maxp augmentation with an improved MobileNetV3 model to classify 11 types of PV defects from IR drone imagery. Their method achieved high F1 scores and demonstrated 30\% faster training over traditional CNNs, making it efficient for field deployment. However, while the model offered significant improvements in training efficiency and defect coverage, its performance dropped in multi-class scenarios with severe class imbalance, and the system still relies on classification without spatial localization capabilities.

Recent trends in embedded AI and smart monitoring have brought attention to lightweight, automated diagnostic systems for PV faults. Ksira et al.\cite{p31} designed a cost-effective embedded setup utilizing TinyCNN on an Arduino microcontroller, capable of detecting common issues such as junction box overheating and dust accumulation from IR images. Their system was integrated with an IoT-enabled cloud platform for remote access. In a parallel effort, Shaban et al.\cite{p32} proposed a comprehensive electroluminescence-based Fault Detection and Classification (FDC) system featuring image preprocessing, feature extraction, feature selection using a modified Butterfly Optimization Algorithm (BOA), and hybrid ensemble classification. While these approaches mark a shift toward intelligent and autonomous PV monitoring, challenges related to scalability, inference latency, and deployment on edge-constrained devices remain.

Conventional inspection techniques, including I–V curve tracing, electroluminescence imaging, and manual visual examination, are often labor-intensive, time-inefficient, or impractical for large-scale solar farms. As a scalable and non-contact diagnostic solution, thermal infrared (TIR) imaging has gained traction due to its ability to reveal surface temperature anomalies indicative of internal defects in PV modules.When combined with unmanned aerial vehicles (UAVs), thermal imaging enables efficient wide-area monitoring of PV arrays. However, interpreting UAV-acquired thermal imagery poses challenges, particularly due to varying lighting conditions, resolution limitations of drone-mounted cameras, noise artifacts, and the non-uniform appearance of different fault types. These limitations restrict the performance of traditional image processing approaches under real-world conditions. To overcome these challenges, recent research has increasingly adopted deep learning-based methods that offer improved robustness, generalization, and spatial precision. While models such as EfficientNet-B0, YOLOv5, and MobileNetV3 have demonstrated strong classification capabilities, many existing solutions still lack pixel-level localization accuracy or are not optimized for real-time deployment on resource-constrained hardware. Furthermore, class imbalance and poor multi-defect separation remain unresolved issues in UAV-based PV fault analysis.

\begin{figure*}[ht!]
\centering
	\includegraphics[width=6cm,height=8.5cm]{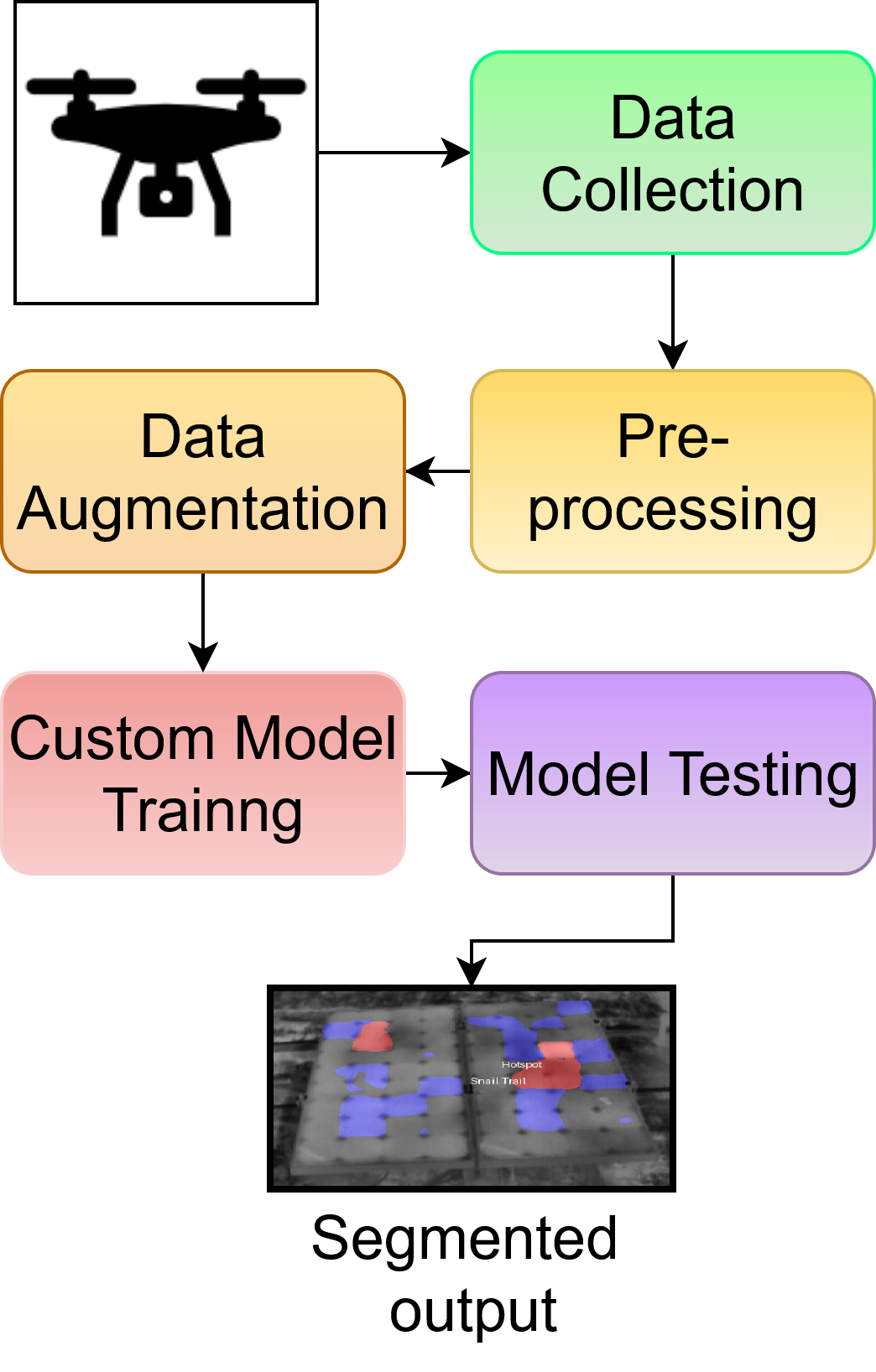}
	\caption{Overview of the proposed methodology.}\label{work}
\end{figure*}

Motivated by these challenges, we present a lightweight deep learning segmentation framework capable of detecting and localizing surface-level PV anomalies, particularly hotspots and snail trails, in aerial thermographic imagery. The proposed approach prioritizes deployment feasibility on embedded systems while ensuring spatial precision and generalization across diverse environmental conditions. The overview of the proposed methodology is shown in Fig. \ref{work}.
The major contributions of this article are summarised as follows:
\begin{itemize}
    \item We propose a customized, lightweight segmentation framework based on the SegFormer architecture, optimized to accurately detect and localize surface-level PV defects such as hotspots and snail trails in aerial thermal imagery
     \item Our methodology demonstrates robust performance under varying environmental conditions, addressing challenges such as resolution variability, noise artifacts, and thermal contrast inconsistencies common in UAV-based PV inspection datasets
     \item We implement a simplified training pipeline that enables pixel-level semantic segmentation without the need for excessive annotation complexity or manual feature engineering.
     \item We benchmark our SegFormer model against multiple state-of-the-art semantic segmentation architectures, including U-Net, DeepLabv3+, PSPNet, and Mask2Former, showing favorable trade-offs in terms of model size, inference time, and segmentation accuracy
     \item The model is designed for deployment feasibility on edge computing platforms, paving the way for real-time PV fault detection in practical UAV-based inspection workflows.
\end{itemize}

\section{Dataset Description}
The dataset used in this study is derived from the publicly available thermal image dataset presented by Alfaro-Mejía et al.\cite{p33}. It comprises a total of 277 thermographic infrared (IR) images captured using a Zenmuse XT radiometric thermal camera mounted on a DJI Matrice 100 UAV platform. The dataset was specifically curated to support the analysis and recognition of two common types of photovoltaic (PV) module surface faults—hotspots and snail trails—in monocrystalline silicon solar panels.

\subsection{Image Acquisition Protocol}
The thermal imagery was acquired during seven aerial inspection sessions conducted over a PV installation site located in Cali, Colombia, under varying environmental conditions. During data collection, the ambient temperature ranged between 26–30 \textdegree C, with wind speeds of 3–5 m/s, and solar irradiance values between 500–1000 W/m². The UAV maintained an approximate 2.3-meter distance from the PV panels to ensure optimal image resolution while minimizing airflow disturbances. Images were captured using the Zenmuse XT camera with a thermal sensitivity of less than 50 mK, operating in the 7.5–13 $\mu$m spectral range, and stored in 336×256 JPEG format.

\begin{figure*}[htbp!]
\centering
\includegraphics[width=0.9\columnwidth]{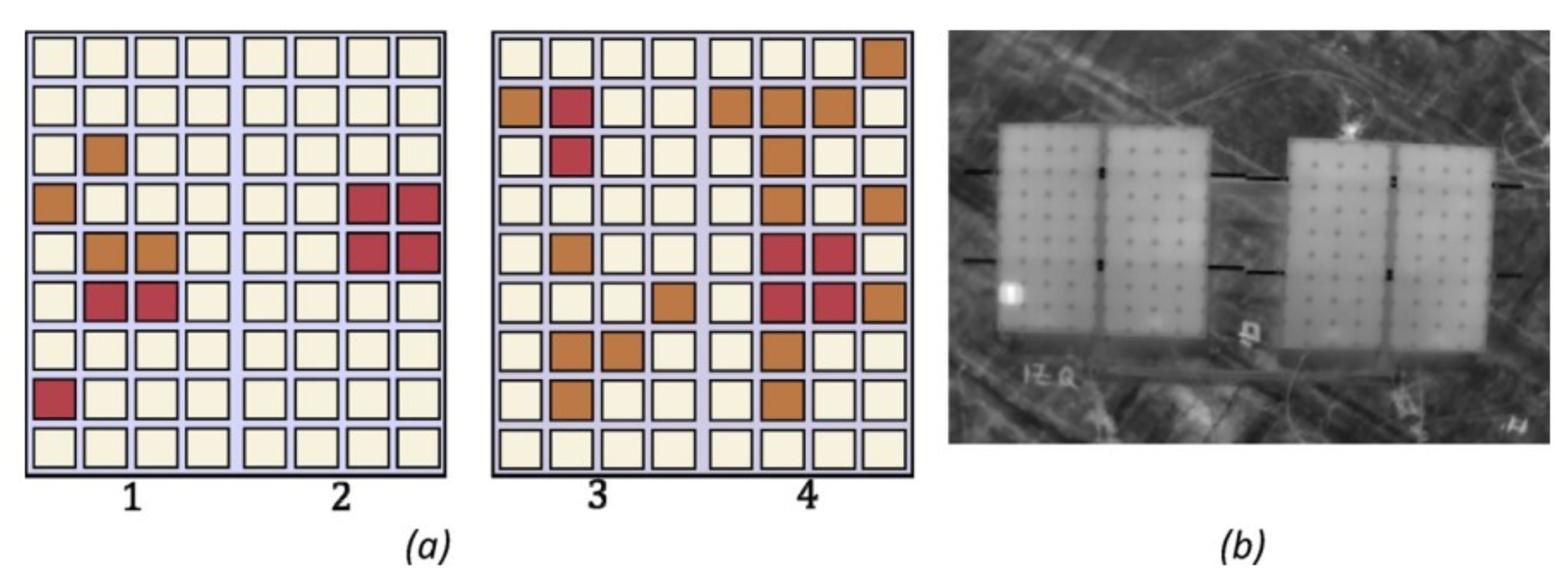}
\caption{Four mono-crystalline Si panels used for experiment: (a) scheme of panels with 9 $\times$ 4 cells in different conditions: hot spot (red) and snail trails (orange) failures, and sound cells (white), labeling of the panels, (b) Thermal image of the real solar panels \cite{p33}.}\label{data}
\end{figure*}

Each flight session was designed to ensure consistent orientation, altitude, and angle of incidence for reliable thermal anomaly detection. The thermal image capture targeted a serial array of four monocrystalline PV panels, with the aim of isolating and analyzing fault patterns at the individual solar cell level. Proper geometric alignment and the instantaneous field of view (IFOV) of the camera ensured that anomalies such as snail trails and hotspots were spatially resolvable within the image frame. The details of hotspots, snailtrails, and good cell is presented in Fig. \ref{data}.

\section{Proposed Methodology}

\subsection{Data Preparation and Preprocessing}

To enable robust and efficient segmentation of defects in photovoltaic (PV) modules, a carefully designed data preprocessing pipeline was implemented. The raw dataset consisted of infrared (IR) thermal images captured using a drone-mounted Zenmuse XT camera, with a native resolution of $336 \times 256$ pixels. Given the limited dataset size and the presence of irrelevant image background, multiple preprocessing steps were performed to enhance signal quality and standardize the input for deep learning models.

\subsubsection{Background Cropping}

The original images include substantial background regions (e.g., vegetation, ground textures) which are thermally and semantically unrelated to the PV modules of interest. These background areas introduce class imbalance and reduce the signal-to-noise ratio, potentially leading the model to learn irrelevant or confounding features. To address this, a region-of-interest (ROI)-based cropping operation was applied to spatially isolate only the PV module areas in each image. This step improves label consistency and ensures that subsequent feature extraction is concentrated on defect-relevant thermal patterns

\subsubsection{Image Resizing}

Following cropping, each image was resized to two different resolutions: the native $336 \times 256$ and a standardized input size of $224 \times 224$. The native resolution preserved the full thermal footprint for defect localization, while the resized version was used to test generalization performance across models trained on conventional image sizes (e.g., EfficientNet input standards). This multi-resolution approach facilitated cross-model evaluation without retraining.

\subsubsection{Contrast Enhancement using CLAHE}

To improve the visibility of subtle thermal anomalies (such as snail trails or small hotspots), we applied Contrast Limited Adaptive Histogram Equalization (CLAHE). Unlike global histogram equalization, CLAHE works on local patches and restricts contrast amplification through clipping:

\begin{equation}
I'(x, y) = \text{CLAHE}(I(x, y), \text{clipLimit}, \text{tileGridSize})
\end{equation}

where $I(x, y)$ is the input image pixel, and $\text{clipLimit} = 2.0$, $\text{tileGridSize} = (8, 8)$ as per our configuration. CLAHE ensures local contrast enhancement while preventing noise over-amplification.

\subsubsection{Denoising using Fast Non-Local Means}

The contrast-enhanced image $I'(x, y)$ was then denoised using the Fast Non-Local Means (NLM) algorithm:

\begin{equation}
I''(x, y) = \text{NLM}(I'(x, y), h, t, s)
\end{equation}

Here, $h = 10$ controls the filtering strength, $t = 7$ is the template window size, and $s = 21$ is the search window size. This denoising step preserves edge information while suppressing noise introduced during image capture or contrast enhancement.

\subsubsection{Min-Max Normalization}

The final step in preprocessing was intensity normalization to a common range $[0, 255]$ using Min-Max scaling. This ensured consistent brightness and contrast distribution across the dataset:

\begin{equation}
I_n(x, y) = \frac{I''(x, y) - I_{\min}}{I_{\max} - I_{\min}} \times 255
\end{equation}

where $I_{\min}$ and $I_{\max}$ are the minimum and maximum pixel intensities in $I''(x, y)$. This step standardizes the dynamic range across all images while preserving thermal intensity variations critical to fault detection.

\begin{figure*}[ht!]
\centering
	\includegraphics[width=6cm,height=8cm]{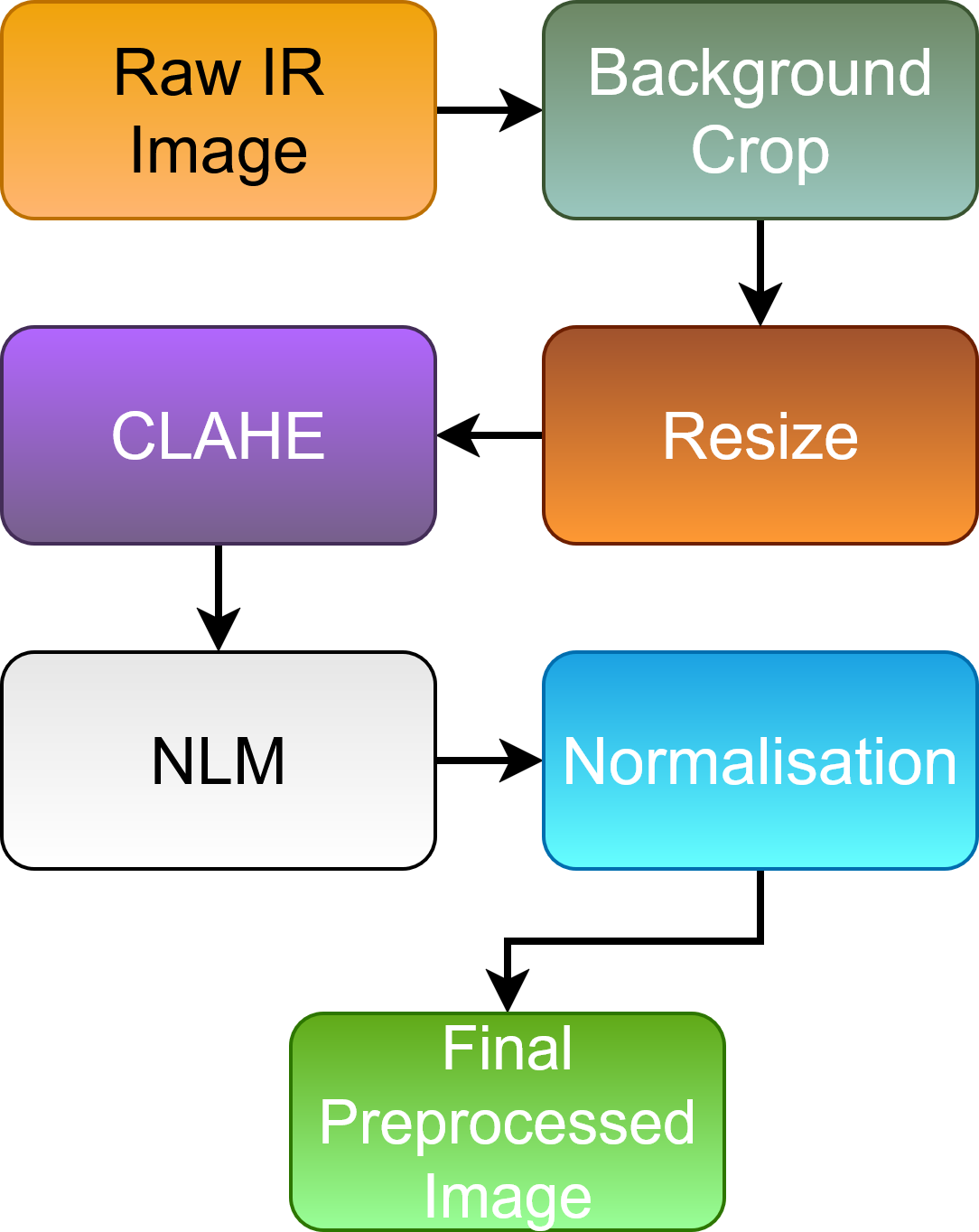}
	\caption{Data preprocessing steps.}\label{pre}
\end{figure*}

The data pre-processing flow is presented in Fig.\ref{pre}. This carefully ordered preprocessing improves thermal contrast, suppresses noise, and standardizes input distributions—ultimately improving model convergence and segmentation accuracy, especially when operating under low-data conditions.

\subsubsection{Data Split Strategy}
To ensure robust evaluation of the segmentation model while maintaining statistical consistency, the dataset was divided into three non-overlapping subsets: training, validation, and testing sets. Out of the total 277 preprocessed thermal images, 221 images were assigned to the training set, 28 images to the validation set, and the remaining 28 images to the test set. This corresponds to an approximate split of 80\% for training, and 10\% each for validation and testing.

This stratified partitioning approach ensures that the model has sufficient data to learn generalizable patterns during training, while also retaining dedicated sets for hyperparameter tuning (validation) and final performance evaluation (testing). Moreover, care was taken to maintain class balance and defect diversity across all subsets, minimizing the risk of data leakage and overfitting during training.

\subsection{Data Augmentation}
\begin{figure*}[ht!]
\centering
	\includegraphics[width=14cm,height=3.5cm]{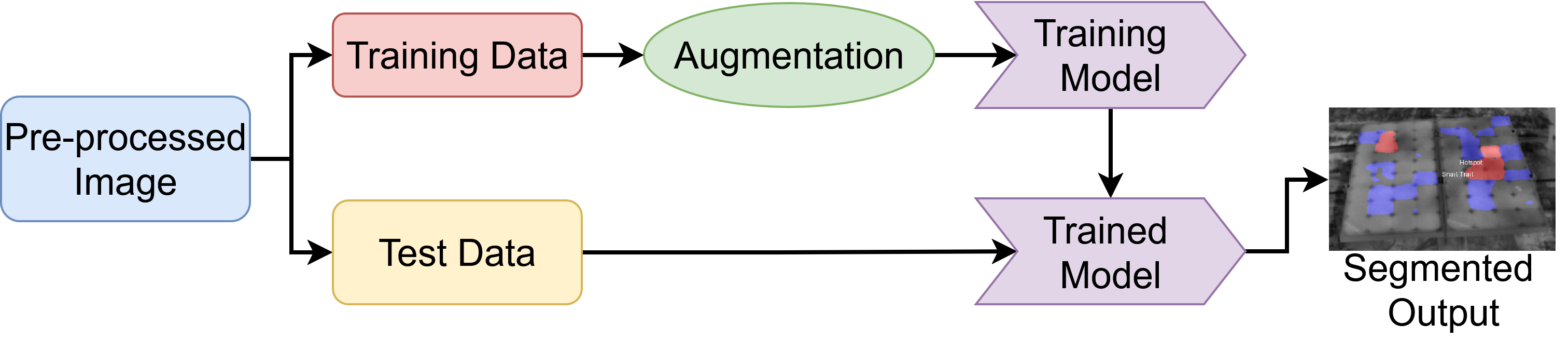}
	\caption{Data Split and Augmentation.}\label{pre2}
\end{figure*}

 To improve generalization and compensate for the limited number of annotated training images, an extensive data augmentation strategy was adopted. This augmentation protocol was designed to simulate realistic variations in environmental conditions, sensor perspectives, and structural appearances of PV modules during UAV-based inspection. Geometric transformations such as horizontal and vertical flips were introduced to emulate the diverse orientations of solar panels encountered during drone flight paths. In addition, random 90-degree rotations and affine transformations—encompassing minor shifts, scaling, and rotation—were applied to encourage the model's invariance to positional and structural distortions. The data split and augmentation step is shown in Fig.\ref{pre2}

Photometric augmentations were incorporated to account for thermal and lighting variability. Specifically, random brightness and contrast adjustments were performed to replicate changes in illumination and sensor response. Gaussian blurring was introduced to mimic focus inconsistencies and moderate resolution loss typical in aerial thermal imaging. To further enhance robustness to occlusions and localized corruptions, coarse spatial dropout was applied, randomly masking small rectangular regions in the image. Moreover, elastic transformations were used to model non-rigid deformations that may occur in PV surfaces due to mechanical stress or installation artifacts.

All augmented images and their corresponding masks were uniformly resized to a spatial resolution of 336$\times$256 pixels to maintain consistency with the model’s input dimensions. Importantly, no augmentation was applied during validation and testing stages to ensure unbiased evaluation of the segmentation performance.

\subsection{Proposed Model}

\subsubsection{Neural Network Foundations}
Artificial neural networks (ANNs) have become foundational tools in modeling high-dimensional, nonlinear data, particularly in computer vision. A neural network is characterized by a set of parameterized transformations:

\begin{equation}
y = f(x; \theta)
\end{equation}

where $x \in \mathbb{R}^n$ is the input, $y \in \mathbb{R}^m$ is the output, and $\theta$ represents learnable weights. Feedforward networks learn hierarchical features through a stack of linear projections followed by non-linear activations such as ReLU, $\tanh$, or $\sigma$. Although powerful, traditional ANNs struggle with capturing long-range dependencies and multi-scale semantics, both of which are critical for pixel-level understanding tasks such as semantic segmentation.

\subsubsection{CNN-Based Segmentation Models}
Convolutional Neural Networks (CNNs) have proven exceptionally effective for dense prediction tasks. In segmentation networks, the encoder–decoder paradigm has become canonical, with an encoder capturing contextual features and a decoder recovering spatial resolution. The segmentation output is generally a pixel-wise class assignment:

\begin{equation}
\hat{y}_{ij} = \arg\max_k \, P(y_{ij} = k \,|\, F_{ij})
\end{equation}

where $F_{ij}$ denotes the feature vector at spatial position $(i,j)$ and $P(y_{ij} = k)$ is derived from a softmax over the logits. Despite their utility, CNNs are inherently limited by local receptive fields, often requiring deep architectures or multi-scale feature aggregation to capture global context.

\subsubsection{Transformer-Based Models for Vision}
The advent of Vision Transformers (ViTs) introduced a paradigm shift by modeling image understanding through self-attention mechanisms. Given an image $X \in \mathbb{R}^{H \times W \times C}$, it is reshaped into a sequence of patches $\{x_1, x_2, \dots, x_N\}$, each embedded and processed as:

\begin{equation}
z_{l} = \mathrm{MSA}(\mathrm{LN}(z_{l-1})) + \mathrm{MLP}(\mathrm{LN}(z_l))
\end{equation}

where $\mathrm{MSA}$ denotes the multi-head self-attention module and $\mathrm{MLP}$ is a feedforward projection. Unlike CNNs, Transformers learn long-range dependencies natively, making them suitable for capturing object-level semantics. However, vanilla ViTs are computationally intensive and lack inherent multi-scale inductive bias.

\subsubsection{SegFormer: A Lightweight and Hierarchical Design}
To reconcile the trade-off between semantic richness and computational efficiency, SegFormer was proposed—a hierarchical Transformer encoder (MiT) coupled with a lightweight all-MLP decoder. The MiT encoder extracts multi-resolution features via overlapping patch embedding and depth-wise convolutions, while the decoder fuses multi-scale context vectors $\{f_1, f_2, f_3, f_4\}$ to predict semantic masks. This hybridization captures both local textures and global semantics without positional embeddings, reducing memory overhead.

\subsubsection{Customised SegFormer for Thermal Image Segmentation}

\begin{figure*}[ht!]
\centering
	\includegraphics[width=14cm,height=3.5cm]{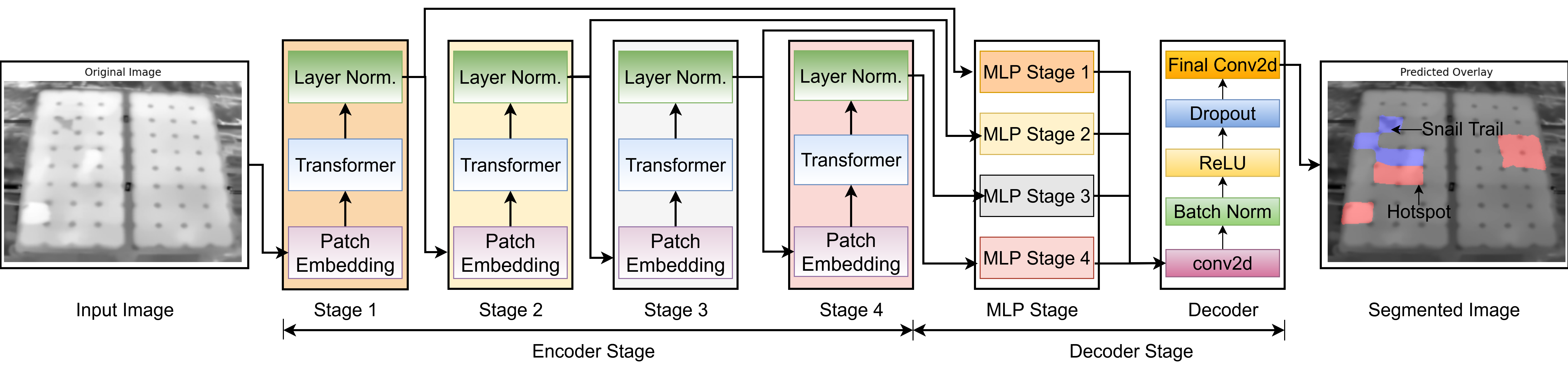}
	\caption{Framework of proposed architecture.}\label{arch}
\end{figure*}

Thermal images of photovoltaic (PV) modules pose specific challenges for semantic segmentation due to their inherently low texture, fuzzy edges, and weak inter-class contrast. These challenges necessitate architectural adaptation to effectively extract and localize defect-relevant features. To this end, we introduce a customized SegFormer-based architecture, restructured to operate under limited training data while ensuring robust feature extraction and efficient computation. The proposed architecture framework is presented in Fig.\ref{arch}

\subsubsection{Encoder Redesign for Multi-Scale Discrimination}
We modify the MiT (Mix Vision Transformer) encoder to emphasize hierarchical feature representation across multiple resolutions. Let the input thermal image be represented as:

\begin{equation}
X \in \mathbb{R}^{H \times W \times 3}
\end{equation}

The image is partitioned into overlapping patches and passed through four transformer stages. Each stage $s \in \{1,2,3,4\}$ produces a feature map:

\begin{equation}
F_s \in \mathbb{R}^{H_s \times W_s \times C_s}
\end{equation}

where $C_s$ is the number of channels and $(H_s, W_s)$ are the downsampled spatial dimensions.

We define the encoder configuration as:
\begin{itemize}
  \item Hidden sizes: $[32, 64, 128, 192]$
  \item Depths per stage: $[2, 2, 3, 3]$
  \item Attention heads: $[1, 2, 4, 6]$
\end{itemize}

Each encoder block follows the transformer formulation:

\begin{equation}
Z_{(l)} = \mathrm{MSA}(\mathrm{LN}(Z_{(l-1)})) + \mathrm{MLP}(\mathrm{LN}(Z_{(l)}))
\end{equation}

where $\mathrm{MSA}$ denotes multi-head self-attention, $\mathrm{LN}$ is layer normalization, and $\mathrm{MLP}$ is a position-wise feed-forward network. These configurations enable progressive feature abstraction while maintaining computational efficiency. The attention mechanism within each stage allows the model to capture long-range dependencies that are crucial for identifying thermal patterns like hotspots and snail trails.

\subsubsection{Decoder Tailoring for Thermal Fault Localization}
To reconstruct fine-grained spatial details from multi-scale encoder outputs $\{F_1, F_2, F_3, F_4\}$, a lightweight MLP-based decoder is employed. The decoder applies a linear projection $\phi_s(\cdot)$ to each feature map:

\begin{equation}
\hat{F}_s = \phi_s(F_s) \in \mathbb{R}^{H' \times W' \times d}
\end{equation}

where $d = 192$ is the decoder hidden size and $H', W'$ are the upsampled spatial dimensions. These projected features are aggregated and upsampled to match the original image resolution.

We define the decoder configuration as follows:
\begin{itemize}
  \item Decoder hidden size: $192$
  \item Upsampling mode: bilinear interpolation
  \item Output convolution: $1 \times 1$ kernel, producing $C = 3$ class logits
\end{itemize}

Final predictions are obtained via a convolutional classifier applied to the concatenated upsampled features:

\begin{equation}
\hat{Y} = \mathrm{Softmax}(\mathrm{Conv}_{1 \times 1}([\hat{F}_1, \hat{F}_2, \hat{F}_3, \hat{F}_4]))
\end{equation}

where $\hat{Y} \in \mathbb{R}^{H \times W \times C}$ is the output probability map over $C = 3$ classes: background, hotspot, and snail trail.

\subsubsection{Domain-Specific Label Encoding}
Given the application domain, the segmentation task is limited to three fault categories. We define a one-hot encoded ground truth tensor:

\begin{equation}
Y \in \{0,1\}^{H \times W \times 3}
\end{equation}

where the third axis corresponds to the class labels. Class weights or focal loss variants can optionally be incorporated to address class imbalance, though we primarily employ categorical cross-entropy.

\subsubsection{Hybrid Transfer Learning Strategy}
To initialize the model, we transfer the encoder weights from the pretrained SegFormer-B0 
The decoder is randomly initialized to accommodate our new label mappings. This hybrid initialization scheme allows us to benefit from large-scale pretraining while adapting the output to thermal-specific defects.

\subsubsection{Training Objective}
The overall segmentation loss is defined as the pixel-averaged categorical cross-entropy:

\begin{equation}
\mathcal{L}_{seg} = -\frac{1}{HW} \sum_{i=1}^{H} \sum_{j=1}^{W} \sum_{c=1}^{C} y_{ij}(c) \log \hat{y}_{ij}(c)
\end{equation}

where $y_{ij}(c)$ and $\hat{y}_{ij}(c)$ are the ground truth and predicted probabilities for class $c$ at pixel $(i,j)$, respectively. This formulation enforces pixel-level discrimination while supporting multi-class predictions. In practice, the softmax operation is applied to logits before computing this loss during training.

\section{Result and Discussion}
In this section, we delve into the details of training, testing, and implementation of the proposed ML model. Apart from these, will also discuss about hyperparameter tuning and evaluate the performance of the proposed model as well as the obtained results.

\subsection{Implementation}

\begin{figure*}[ht!]
\centering
	\includegraphics[width=14cm,height=10cm]{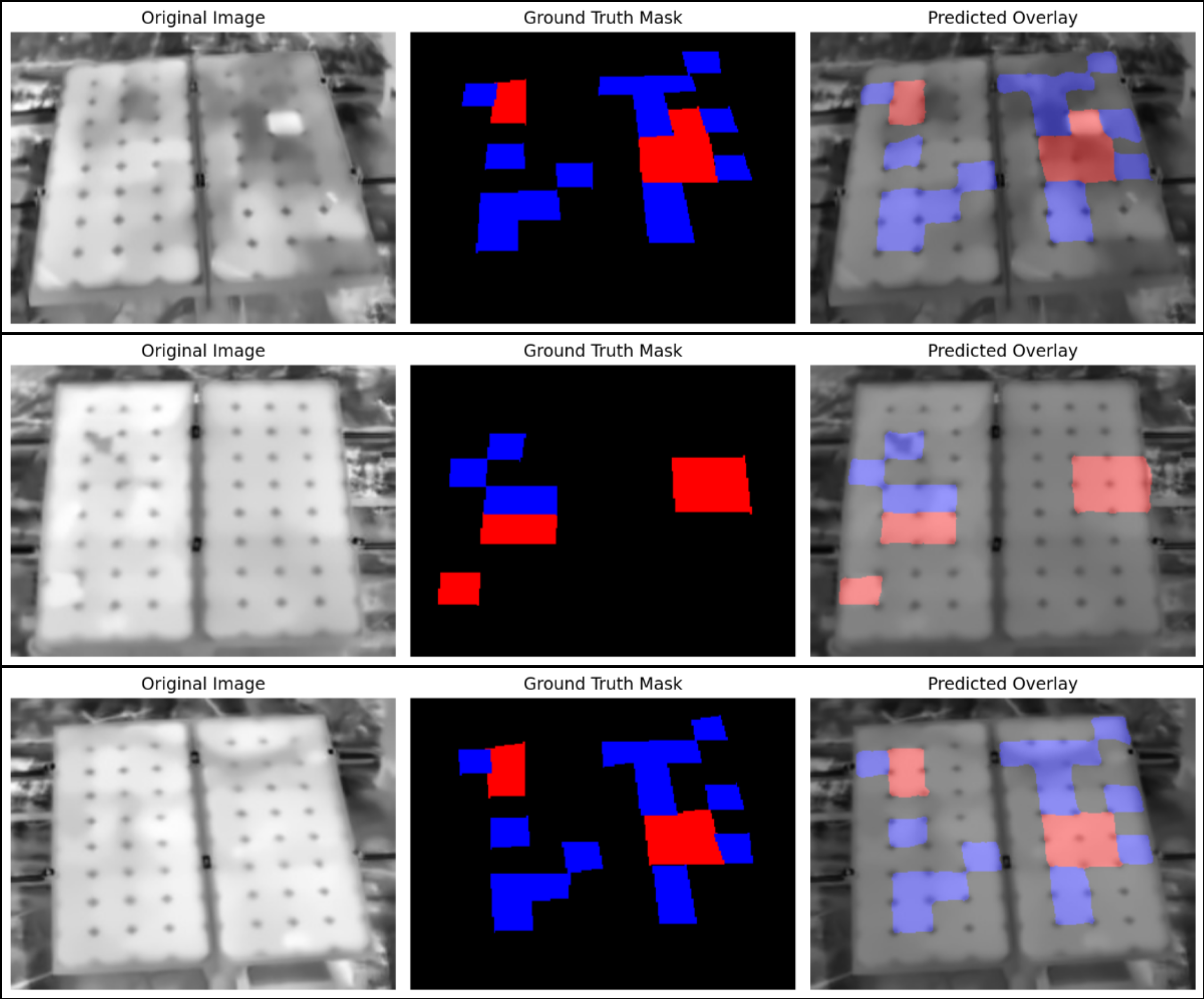}
	\caption{Segmented hotspot and snail trail from proposed custom model.}\label{final}
\end{figure*}

The model was developed using Python version 3.10.16, a widely adopted programming language for deep learning applications due to its extensive ecosystem and flexibility. Model implementation was carried out using PyTorch version 2.5.1 with CUDA 12.1 support, enabling efficient utilization of GPU resources during training. The training process was accelerated using an NVIDIA RTX 1000 Ada Generation GPU, which provided the necessary computational power to handle the high-dimensional operations intrinsic to deep neural network training. This GPU acceleration significantly reduced model convergence time and improved training efficiency. All computations were performed on a Windows 11-based 64-bit system equipped with an Intel Core Ultra 7 165H processor running at 1.40 GHz and 32 GB of RAM.
The final segmented output of the images from the custom model is shown in Fig.\ref{final}. In Fig.\ref{final} red colour segmentation is for hotspot and blue colour is for snailtrail.

\subsection{Hyperparameter Tuning}

\begin{table}[ht!]
\centering
\caption{Architecture of the Proposed Model}
\begin{tabular}{|c|c|c|}
\hline
\textbf{Layer Type} & \textbf{Output Shape} & \textbf{\begin{tabular}[c]{@{}c@{}}Total Parameter\\ (Weights + Biases)\end{tabular}} \\ \hline
Input               & 3x256x336             & 0                                                                                     \\ \hline
Stage 1 PatchEmb    & 5376x32               & 4,800                                                                                 \\ \hline
Stage 1 Transformer & 5376x32               & 159,232                                                                               \\ \hline
Stage 1 LayerNorm   & 5376x32               & 64                                                                                    \\ \hline
Stage 2 PatchEmb    & 1344x64               & 18,624                                                                                \\ \hline
Stage 2 Transformer & 1344x64               & 236,544                                                                               \\ \hline
Stage 2 LayerNorm   & 1344x64               & 128                                                                                   \\ \hline
Stage 3 PatchEmb    & 336x128               & 74,112                                                                                \\ \hline
Stage 3 Transformer & 336x128               & 807,936                                                                               \\ \hline
Stage 3 LayerNorm   & 336x128               & 256                                                                                   \\ \hline
Stage 4 PatchEmb    & 88x192                & 221,760                                                                               \\ \hline
Stage 4 Transformer & 88x192                & 1,357,632                                                                             \\ \hline
Stage 4 LayerNorm   & 88x192                & 384                                                                                   \\ \hline
MLP Stage 1         & 5376x192              & 6,336                                                                                 \\ \hline
MLP Stage 2         & 1344x192              & 12,480                                                                                \\ \hline
MLP Stage 3         & 336x192               & 24,768                                                                                \\ \hline
MLP Stage 4         & 88x192                & 37,056                                                                                \\ \hline
Conv2d              & 192x64x84             & 147,456                                                                               \\ \hline
BatchNorm           & 192x64x84             & 384                                                                                   \\ \hline
ReLU                & 192x64x84             & 0                                                                                     \\ \hline
Dropout             & 192x64x84             & 0                                                                                     \\ \hline
Final Conv2d        & 3x64x84               & 579                                                                                   \\ \hline
Output              & 3x64x84               & 0                                                                                     \\ \hline
\end{tabular}
\label{table_segformer}
\end{table}
Hyperparameter tuning plays a pivotal role in optimizing the performance of deep learning models, particularly in scenarios with limited domain-specific data such as thermal photovoltaic (PV) imagery. These hyperparameters—unlike trainable parameters—must be defined a priori and significantly influence convergence behavior, model generalization, and training stability. In this work, we employed a grid search strategy to determine optimal values for key hyperparameters including learning rate, batch size, encoder-decoder dimensions, and activation functions.

The learning rate, a critical parameter controlling the step size of gradient updates, was varied from 1e-2 to 1e-5. After iterative experimentation, a value of 1e-4 was found to offer a stable trade-off between convergence speed and final segmentation accuracy. Batch size, which governs the number of training samples per update, was explored in the range of 2 to 16. A batch size of 4 was selected as it provided the best stability without compromising GPU memory constraints or convergence consistency.

The hierarchical encoder of our customized SegFormer model comprises four stages, with hidden sizes set to [32, 64, 128, 192] and depths configured as [2, 2, 3, 3]. These settings were fine-tuned to preserve computational efficiency while maximizing semantic abstraction at multiple scales. The decoder hidden size was also evaluated at multiple levels (128, 192, 256), and a value of 192 was finalized as it enabled effective multi-scale feature fusion without over-parameterization.

To investigate the impact of nonlinearity, we tested multiple activation functions including ReLU, GELU, and SiLU across intermediate layers. The default GELU activation in transformer blocks was retained, as it provided smooth convergence in conjunction with the self-attention mechanism. Dropout rates within the classifier were tested between 0.05 and 0.2, with 0.1 offering the best generalization.

All hyperparameter combinations were evaluated using a validation set, with pixel-wise cross-entropy loss and mean Intersection-over-Union (mIoU) serving as the primary performance metrics. The tuning process followed a multi-round iterative refinement protocol, progressively narrowing the search space based on empirical trends. This meticulous optimization yielded the final model configuration, which demonstrated improved segmentation performance while remaining lightweight and deployment-ready for real-world PV inspection scenarios, final hyperparameters which are used in custom model is presented in Table \ref{table_segformer}.

\subsection{Quantitative Evaluation Metrics}

The performance of the proposed customized SegFormer model is evaluated using a set of standard quantitative metrics. These include average test loss, mean Intersection-over-Union (mIoU), pixel accuracy, F1 score, precision, recall, Cohen’s kappa score, and per-class Dice coefficients.

The model achieves an average test loss of 0.0844, indicating stable convergence. The mean IoU is 83.94\%, reflecting strong spatial overlap between predicted and ground truth masks. A high pixel accuracy of 97.65\% confirms robust per-pixel classification across background, hotspot, and snail trail regions.

Evaluation across individual classes yields a macro F1-score of 94.47\%, average precision of 93.15\%, and average recall of 93.79\%, indicating consistent performance across all classes. The Cohen’s kappa score of 0.9052, further demonstrates excellent agreement between predicted and actual segmentation, even when accounting for random chance.

Per-class Dice coefficients are computed as: 0.9873 for Background, 0.9032 for Hotspot, and 0.9135 for Snail Trail, validating high segmentation accuracy for each category. The evaluation values presented here are considered after applying k-fold cross validation where value of $k$ is 5.

Table~\ref{tab:eval_metrics} summarizes the detailed classification performance, including class-wise precision, recall, and F1-scores on the test set.

\begin{table}[ht!]
\centering
\caption{Detailed classification performance on test dataset for the proposed customized SegFormer model.}
\begin{tabular}{|l|l|l|l|}
\hline
\textbf{Class} & \textbf{Precision} & \textbf{Recall} & \textbf{F1-score} \\ \hline
Background     & 98.81\%            & 98.65\%         & 98.73\%           \\
Hotspot        & 89.31\%            & 91.36\%         & 90.32\%           \\
Snail Trail    & 91.33\%            & 91.37\%         & 91.35\%           \\ \hline
Macro Avg.     & 93.15\%            & 93.79\%         & 93.47\%           \\
Weighted Avg.  & 97.66\%            & 97.65\%         & 97.66\%           \\ \hline
\end{tabular}
\label{tab:eval_metrics}
\end{table}

\subsection{Comparative Study}
In this work, we present a comprehensive evaluation of our customized SegFormer model, specifically designed for semantic segmentation of thermal images in photovoltaic (PV) fault detection. The comparative study is structured across three distinct dimensions to holistically assess the architectural efficacy and generalization capability of the proposed model. First, we examine performance variations across multiple SegFormer variants (B0–B4) by altering encoder depths, hidden dimensions, and attention head configurations, thereby evaluating architectural scalability and efficiency. Second, we benchmark our model against a suite of other standard deep learning-based segmentation frameworks, such as U-Net, DeepLabv3, PSPNet, Mask2Former. Finally, we assess the effectiveness of our customized SegFormer model relative to segmentation methods reported in the literature . This includes benchmarking against state-of-the-art architectures applied to similar thermal imaging tasks in photovoltaic fault detection. Through this tripartite evaluation framework, we aim to rigorously analyze the proposed model’s scalability, generalization ability, and robustness across varying architectural designs, comparative baselines, and domain-relevant segmentation challenges.

\subsubsection{Segformer Model Type}

\begin{table}[ht]
\centering
\caption{Comparison of SegFormer model variants}
\begin{tabular}{|l|l|l|l|l|l|}
\hline
\textbf{Model}    & \textbf{\begin{tabular}[c]{@{}l@{}}Model\\ Size (MB)\end{tabular}} & \textbf{Parameters} & \textbf{\begin{tabular}[c]{@{}l@{}}Avg.\\ loss\end{tabular}} & \textbf{\begin{tabular}[c]{@{}l@{}}Avg.\\ IoU\end{tabular}} & \textbf{\begin{tabular}[c]{@{}l@{}}Pixel\\ Accuracy\end{tabular}} \\ \hline
B0                & 132                                                                & 3.7M                & 0.3212                                                       & 0.8443                                                      & 0.9583                                                            \\
B1                & 251                                                                & 13.6M               & 0.3034                                                       & 0.8524                                                      & 0.9596                                                            \\
B2                & 505                                                                & 27.3M               & 0.3403                                                       & 0.8599                                                      & 0.9612                                                            \\
B3                & 729                                                                & 47.3M               & 0.3618                                                       & 0.8575                                                      & 0.9608                                                            \\
B4                & 976                                                                & 63.9M               & 0.3411                                                       & 0.8567                                                      & 0.9605                                                            \\
\textbf{Proposed} & \textbf{123}                                                       & \textbf{3.1M}       & \textbf{0.3042}                                              & \textbf{0.8263}                                             & \textbf{0.9546}                                                   \\ \hline
\end{tabular}
\label{tab:segformer_models}
\end{table}
To evaluate the architectural efficiency of our customized model, we conducted a comparative analysis across various variants of the SegFormer architecture, namely B0 through B4. Each variant is characterized by increasing encoder depth, attention complexity, and parameter count \cite{p37}. As model complexity increases, so do the computational and memory requirements, making it critical to assess the trade-offs between performance and efficiency. The comparison is shown in Table \ref{tab:segformer_models}.
The B0 model, being the lightest with 3.7 million parameters and a model size of 132MB, achieved a mean Intersection over Union (IoU) of 0.8443 and a pixel accuracy of 95.83\%. Performance improved with the B1 and B2 variants, with B2 achieving the highest average IoU of 0.8599 and pixel accuracy of 96.12\%, albeit with a significantly larger model footprint (505MB and 27.3 million parameters). Further increasing complexity with B3 and B4 did not yield substantial gains, and in some cases, even resulted in marginal degradation in segmentation performance, indicating diminishing returns.

Our proposed customized SegFormer strikes a balance between efficiency and accuracy. With only 3.1 million parameters and a model size of 123MB, it maintains a competitive average loss (0.3042), IoU (0.8263), and pixel accuracy (95.46\%). This demonstrates that architectural tailoring can retain performance while significantly reducing computational overhead, making the model suitable for real-time and embedded deployment.

\subsubsection{Different Standard Models}
\begin{table}[htbp]
\centering
\caption{Performance Comparison with Different Standard Models.}
\begin{tabular}{|l|l|l|l|l|l|}
\hline
\textbf{Model}                                                    & \textbf{\begin{tabular}[c]{@{}l@{}}Model\\ Size(MB)\end{tabular}} & \textbf{Parameters} & \textbf{\begin{tabular}[c]{@{}l@{}}Avg.\\ Loss\end{tabular}} & \textbf{\begin{tabular}[c]{@{}l@{}}Avg.\\ IoU\end{tabular}} & \textbf{\begin{tabular}[c]{@{}l@{}}Pixel\\ Accuracy\end{tabular}} \\ \hline
U-net                                                             & 1460                                                              & 31M                 & 0.2241                                                       & 0.7565                                                      & 0.9441                                                            \\
\begin{tabular}[c]{@{}l@{}}DeepLab-\\ v3\end{tabular}             & 1212                                                              & 23.5M               & 0.1834                                                       & 0.8015                                                      & 0.9551                                                            \\
PSPNet                                                            & 850                                                               & 47M                 & 0.3354                                                       & 0.3582                                                      & 0.7707                                                            \\
\begin{tabular}[c]{@{}l@{}}Mask2\\ Former\end{tabular}            & 750                                                               & 43M                 & 0.3618                                                       & 0.6233                                                      & 0.7308                                                            \\
\textbf{\begin{tabular}[c]{@{}l@{}}Proposed\\ Model\end{tabular}} & \textbf{123}                                                      & \textbf{3.1M}       & \textbf{0.3042}                                              & \textbf{0.8263}                                             & \textbf{0.9546}                                                   \\ \hline
\end{tabular}
\label{tab:standard_models}
\end{table}
To further assess the effectiveness of the proposed customized SegFormer model, we compare it with several widely adopted semantic segmentation architectures across the computer vision domain. These include U-Net\cite{p38}, DeepLabv3\cite{p39}, PSPNet\cite{p40}, and Mask2Former\cite{p41}. Each of these architectures has distinct design philosophies and complexities that influence their suitability for thermal image segmentation tasks in photovoltaic (PV) modules. The comparison is shown in Table \ref{tab:standard_models}.

U-Net, with its encoder–decoder structure and skip connections, is a popular architecture for medical and infrared image segmentation due to its spatial localization capabilities. However, its large model size (1460 MB) and 31 million parameters make it computationally expensive, especially for edge deployments. DeepLabv3, which integrates atrous spatial pyramid pooling (ASPP), achieves high pixel accuracy (95.51\%) and improved IoU (80.15\%) but still demands over 1.2 GB in memory.

PSPNet incorporates pyramid pooling to capture global context. Despite having 47 million parameters, it significantly underperforms on our dataset with a mean IoU of only 35.82\% and pixel accuracy of 77.07\%. Mask2Former, a transformer-based mask prediction framework, also struggles with thermal image segmentation, yielding an IoU of 62.33\% and the highest test loss among all evaluated models (0.3618).

In contrast, our customized SegFormer model strikes an efficient balance between compactness and performance. It achieves an IoU of 82.63\% and pixel accuracy of 95.46\% with only 3.1 million parameters and a 123 MB model size—significantly smaller than other models. This demonstrates the proposed model’s suitability for lightweight, real-time segmentation tasks under resource-constrained environments without compromising segmentation quality.

\subsubsection{Comparison with Other Methods }
To evaluate the broader applicability and effectiveness of our customized SegFormer model, we compare it with several recent state-of-the-art techniques for photovoltaic (PV) fault detection that use different segmentation strategies, data modalities, and architectural principles. This comparison provides insight into the strengths and limitations of each approach and contextualizes the improvements introduced by our method.

In the study by Mahboob et al. \cite{p34}, a standard SegFormer was employed for semantic cell segmentation on electroluminescence (EL) images, achieving a pixel accuracy of 96.2\% and a mean IoU of 56.5\%. While this demonstrates the potential of transformer-based architectures, the dataset used predominantly features microcracks and cell-level defects captured using EL imaging. In contrast, our proposed model targets more complex surface-level anomalies such as hotspots and snail trails in real-world field-acquired thermal infrared (IR) images, which are inherently noisier and present greater challenges for segmentation. Despite these difficulties, our model achieved a comparable pixel accuracy of 95.46\% and a significantly higher mean IoU of 82.63\%, indicating improved robustness and generalization in practical PV inspection scenarios.

Another approach using classical signal processing and rough set theory applied to UAV-acquired RGB images achieved a classification accuracy of 89.68\% but did not support segmentation \cite{p13}. While suitable for coarse-level fault detection, this method lacks the ability to localize specific fault regions, which our approach handles efficiently through dense semantic segmentation.

A dual-stage M-shaped network was proposed for combined classification and segmentation of microcracks\cite{p27}. It achieved a mean IoU of 0.693 on a public microcrack dataset but was limited to binary segmentation. Our model not only supports multi-class segmentation (e.g., hotspots and snail trails) but also adapts to thermal imaging, offering broader applicability.

Lastly, an enhanced U-Net using EL polarization imaging achieved high accuracy (92.66\% pixel accuracy and 86.05\% mean IoU) for segmenting microcracks \cite{p26}. However, this method was designed for a single class in a controlled lab setting. In contrast, our model targets three fault classes under real environmental conditions using low-cost thermal imagery, making it more deployable for on-field inspection.

In summary, our customized SegFormer model demonstrates a superior balance of accuracy, generalizability, and efficiency across varying image modalities and segmentation complexity. The comparison validates the model’s capability to handle diverse input characteristics and offers practical relevance for real-world solar panel monitoring systems. The comparison with other existing work is presented in Table \ref{tab:method_comparison}.

\begin{table}[ht]
\caption{Comparison of the proposed customized SegFormer with existing methods for PV module fault detection.}
\begin{tabular}{|l|l|l|l|l|l|l|l|}
\hline
Paper    & \begin{tabular}[c]{@{}l@{}}Method/\\ Model\end{tabular}                  & \begin{tabular}[c]{@{}l@{}}Segmen-\\ tation\end{tabular} & \begin{tabular}[c]{@{}l@{}}No. of\\ Classes\end{tabular} & \begin{tabular}[c]{@{}l@{}}Dataset\\ Type\end{tabular}                          & \begin{tabular}[c]{@{}l@{}}Pixel\\ Accuracy\end{tabular} & \begin{tabular}[c]{@{}l@{}}Mean\\ IoU\end{tabular} & Loss   \\ \hline
\cite{p34}       & \begin{tabular}[c]{@{}l@{}}Standard\\ Segformer\end{tabular}             & Yes                                                      & 6                                                        & \begin{tabular}[c]{@{}l@{}}Public\\ EL images\\ \cite{p35}\end{tabular}                 & 96.2\%                                                   & 56.5\%                                             & NR     \\
\cite{p13}       & \begin{tabular}[c]{@{}l@{}}Classical\\ Signal \\ Processing\end{tabular} & No                                                       & 5                                                        & \begin{tabular}[c]{@{}l@{}}Private\\ RGB Images\end{tabular}                    & 89.68\%                                                  & NR                                                 & NR     \\
\cite{p27}       & \begin{tabular}[c]{@{}l@{}}2-Stage\\ M-Shape \\ N/W\end{tabular}         & Yes                                                      & 2                                                        & \begin{tabular}[c]{@{}l@{}}Public \\ EL-images\\ \cite{p36}\end{tabular}                & NR                                                       & 69.3\%                                             & NR     \\
\cite{p26}       & \begin{tabular}[c]{@{}l@{}}Enhanced\\  U-Net\end{tabular}                & Yes                                                      & 1                                                        & \begin{tabular}[c]{@{}l@{}}Lab -based\\ EL images\end{tabular}                  & 92.66\%                                                  & 86.05\%                                            & NR     \\
Proposed & \begin{tabular}[c]{@{}l@{}}Custom Low\\ Complex\\ Segformer\end{tabular} & Yes                                                      & 3                                                        & \begin{tabular}[c]{@{}l@{}}Public\\ IR Images\\ \cite{p33}\end{tabular} & 95.46\%                                                  & 82.63\%                                            & 0.3042 \\ \hline
\end{tabular}
\label{tab:method_comparison}
\end{table}

\section{Conclusion}
In this study, we presented a customized SegFormer-based semantic segmentation model tailored specifically for identifying surface-level faults—namely hotspots and snail trails—in photovoltaic (PV) modules using low-cost thermal imagery. Unlike prior works that rely heavily on electroluminescence or RGB imaging captured under controlled laboratory settings, our approach addresses the challenges posed by noisy, real-world thermal images acquired in uncontrolled environments. Through architectural simplifications and domain-specific decoder modifications, our lightweight model (3.1M parameters, 123MB size) achieves high segmentation performance while maintaining computational efficiency, making it suitable for deployment in edge or resource-constrained systems.

Extensive experiments demonstrate that the proposed model not only outperforms classical machine learning baselines and standalone CNNs but also exhibits competitive or superior performance compared to larger transformer-based variants (e.g., SegFormer-B1 to B4) and well-established segmentation frameworks such as U-Net, DeepLabV3, PSPNet, and Mask2Former. Our model achieved a mean IoU of 82.63\% and a pixel accuracy of 95.46\%, while requiring fewer computational resources and achieving faster inference times.

Furthermore, our comparative evaluation against state-of-the-art methods in recent literature highlights the advantages of our model in terms of generalizability, multi-class segmentation capability, and adaptability to thermal domain inputs. Unlike conventional models limited to single-class microcrack detection or binary segmentation, our method robustly segments multiple fault types, providing localized diagnostics necessary for automated maintenance workflows in solar farms.

In addition to achieving high accuracy and efficiency, the use of a publicly available dataset, rigorous experimental setup, and a lightweight transformer-based architecture enhances the practical relevance of the proposed model for real-world deployment. Future research directions include expanding the model’s generalizability across varying environmental conditions, integrating temporal dynamics through video-based drone inspections, and incorporating uncertainty quantification for assessing fault severity. Overall, this work marks a significant advancement toward robust, scalable, and deployable PV fault detection systems based on transformer-driven semantic segmentation.








  \bibliographystyle{elsarticle-num} 
  \bibliography{Main_file}





\end{document}